\documentclass{IEEEtran}
\usepackage{xcolor}
\usepackage{graphicx}  
\usepackage{generic}
\usepackage{cite}
\usepackage{amsmath,amssymb,amsfonts}
\usepackage[utf8]{inputenc}
\usepackage{graphicx}  
\usepackage{rotating}  
\usepackage{longtable} 

\usepackage{hyperref}
\usepackage{textcomp}
\usepackage{booktabs}
\usepackage{stfloats}
\usepackage[T1]{fontenc}
\usepackage{lipsum}  
\usepackage{enumitem}  
\usepackage{algpseudocode} 
\usepackage{array}
\usepackage{generic}

\usepackage[ruled]{algorithm}

\begin{document}

\title{Advanced Multimodal Learning for Seizure Detection and Prediction: Concept, Challenges, and Future Directions }
\author{Ijaz Ahmad, Faizan Ahmad, Sunday Timothy Aboyeji,  Yongtao Zhang, Peng Yang, Javed Ali Khan, Rab Nawaz,  Baiying Lei
\thanks{This work was supported partly by Shenzhen Science and Technology Program (Grant Nos. JCYJ20220530142405013,4202004 and JCYJ20220818095809021) and Shenzhen Medical Research Funds (Nos.C2401023 and C2301005).(Corresponding author: Baiying Lei ). }
\thanks{I.Ahmad, Y.Zhange and P.Yang is with the National-Regional Key Technology Engineering Laboratory for Medical Ultrasound, Guangdong Key Laboratory for Biomedical Measurements and Ultrasound Imaging, School of Biomedical Engineering, Shenzhen University Medical School, Marshall Laboratory of Biomedical Engineering, Shenzhen University, Shenzhen, 518060, China(Email:ijaz@szu.edu.cn, zhangyt@szu.edu.cn,peng.yang@szu.edu.cn ) }
\thanks{F.Ahmad is with the College of Big Data and Internet, Shenzhen Technology University, Shenzhen, China. (Email:ahmadfaizan@sztu.edu.cn).}
\thanks{S.T.Aboyeji is S. T. Aboyeji is with the Department of Otolaryngology, Shenzhen Longgang Otolaryngology Hospital, and also with Shenzhen Otolaryngology Research Institute, Shenzhen, China.  China.(Email: sundayaboyeji@ieee.org).}
\thanks{R.Nawaz is with the School of Computer Science and Electronic Engineering, University of Essex, Colchester, United Kingdom (Email:rab.nawaz@essex.ac.uk)}
\thanks{B.Lei is with the National-Regional Key Technology Engineering Laboratory for Medical Ultrasound, Guangdong Key Laboratory for Biomedical Measurements and Ultrasound Imaging, School of Biomedical Engineering, Shenzhen University Medical School, South China Hospital, Shenzhen University, Shenzhen 518055, China (Email: leiby@szu.edu.cn; zhangtd270@163.com; qujunlong2022@email.szu.edu.cn; 1104366561@qq.com ).}
}
\maketitle

\begin{abstract}
Epilepsy is a chronic neurological disorder characterized by recurrent unprovoked seizures, affects over 50 million people worldwide, and poses significant risks, including sudden unexpected death in epilepsy (SUDEP). Conventional unimodal approaches, primarily reliant on electroencephalography (EEG), face several key challenges, including low SNR, nonstationarity, inter- and intrapatient heterogeneity, portability, and real-time applicability in clinical settings. To address these issues, a comprehensive survey highlights the concept of advanced multimodal learning for epileptic seizure detection and prediction (AMLSDP). The survey presents the evolution of epileptic seizure detection (ESD) and prediction (ESP) technologies across different eras. The survey also explores the core challenges of multimodal and non-EEG-based ESD and ESP. To overcome the key challenges of the multimodal system, the survey introduces the advanced processing strategies for efficient AMLSDP. Furthermore, this survey highlights future directions for researchers and practitioners. We believe this work will advance neurotechnology toward wearable and imaging-based solutions for epilepsy monitoring, serving as a valuable resource for future innovations in this domain.
\end{abstract}

\begin{IEEEkeywords}

Epilepsy, Multimodal Learning, Seizure Detection

\end{IEEEkeywords}

\section{Introduction}
\begin{figure*}[!ht]
 \centering
\includegraphics[width=\linewidth]{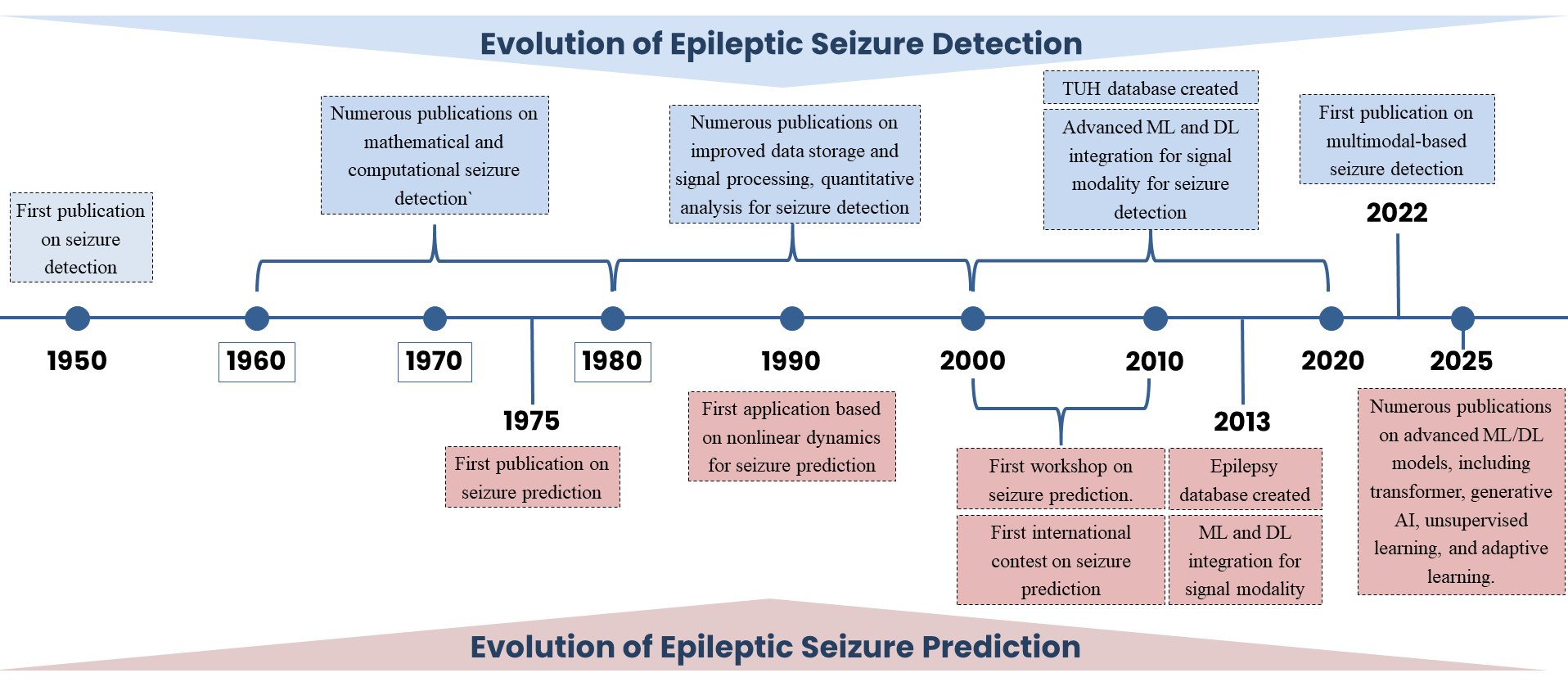}
    \centering
   \caption{ Evolution of epilepsy management technologies, showing seizure detection (1950-2025) and prediction (1975-2025) timelines. The flow highlights the shift in the evaluation of the biomedical signals processing and key AI milestones driving real-time epileptic  seizure detection and prediction.}
  \label{fig:2}
 \end{figure*}

Epilepsy is a group of neurological disorders characterised by an enduring predisposition to generate recurrent, unprovoked seizures arising from abnormal electrical activity in the brain \cite{sr1}. Affecting approximately 50 million individuals worldwide, it poses substantial challenges to quality of life, including risks of injury, social stigma, and comorbidities such as cognitive impairments and psychiatric conditions \cite{WHO_Epilepsy_2024}. The disorder can originate from genetic, structural, or idiopathic causes, underscoring its status as a major public health concern. Early diagnosis and personalized treatment are essential to mitigate long-term impacts, necessitating advanced tools for identifying epileptic activity and localizing affected brain regions.
\begin{figure*}[!ht]
 \centering
\includegraphics[width=7.5in, height=4in]{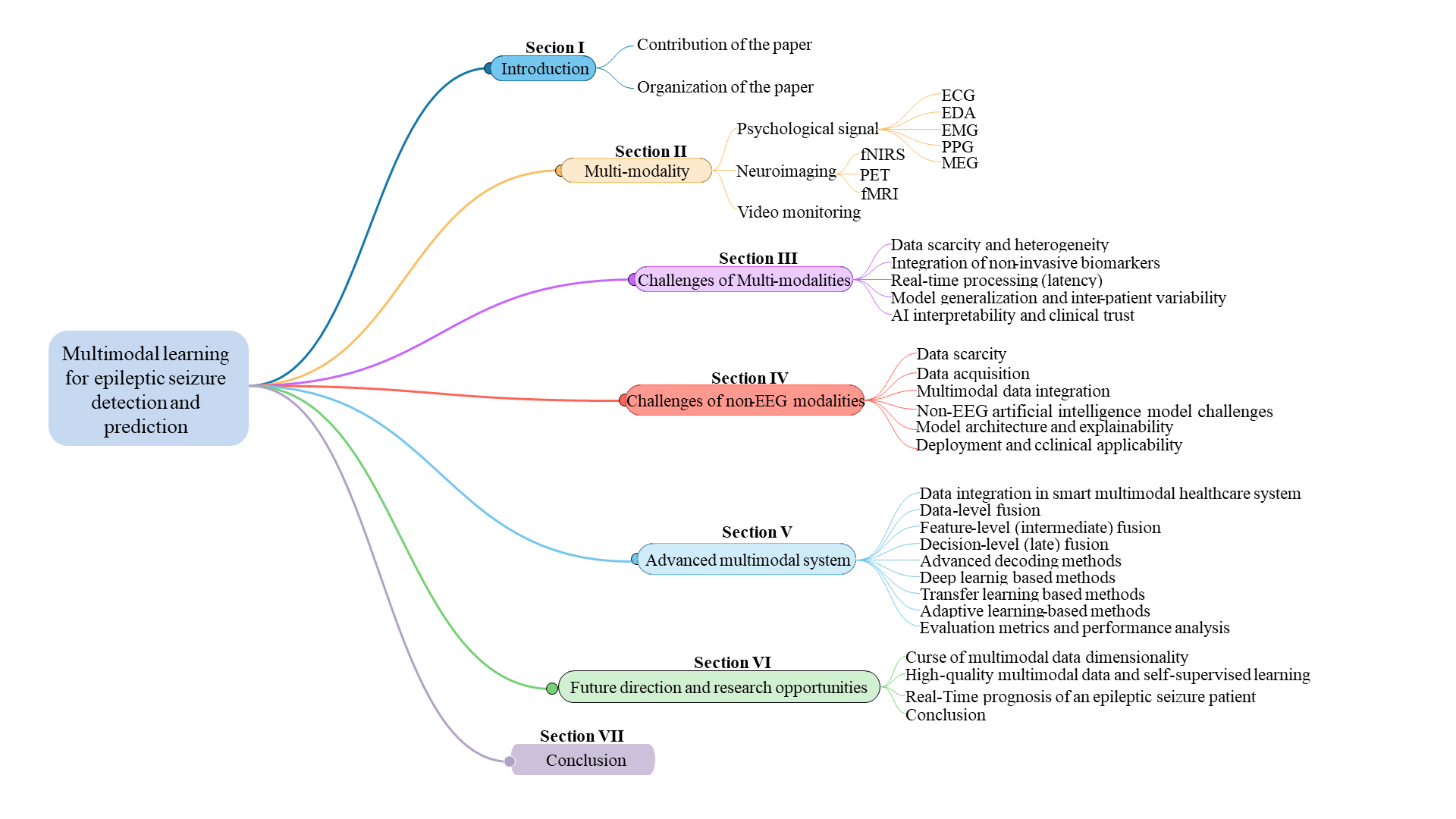}
    \centering
  \caption{Outline of the survey paper. We explore a comprehensive overview of advanced multimodal learning for ESD and ESP, including the multimodal monitoring system, the core challenges of non-EEG and multimodality-based seizure detection and prediction, advanced strategies, and future directions.}
  \label{fig:1}
 \end{figure*}
Electroencephalography (EEG) serves as the primary tool for seizure diagnosis due to its widespread availability, affordability, and suitability for outpatient monitoring. Ambulatory EEG (aEEG) extends this capability by enabling long-term recordings in naturalistic environments while maintaining patient mobility \cite{sr4}.The discovery of human EEG by Hans Berger in the late 1920s initially relied on visual inspection for analysis \cite{gloor1969hans}. As shown in Fig. \ref{fig:2} from 1960s and 1970s marked a pivotal shift toward mathematical and computational modeling, driven by pioneers such as Herbert Vaughan Jr., Derek Fender, and Dietrich Lehmann \cite{sr5}. Since the 1970s, epileptic seizure detection (ESD) and prediction (ESP) have progressed through distinct eras of signal processing and machine learning advancements. Early efforts (1970s to mid-1980s) focused on automatic classification via spike and transient detection, employing spectral analysis, time-frequency methods, wavelet transforms, and rudimentary neural networks \cite{sr6,sr7,sr8,sr9}. The similar works included ESD using implanted electrodes and custom circuitry \cite{sr11}, decomposition of EEG signals into elementary waves for classifying paroxysmal bursts \cite{sr12}, and quantitative analyses of spiking rates correlated with seizures and pharmacological levels \cite{sr6,sr10}. Rule-based approaches, such as thresholds on rhythmic bursts, amplitude, and frequency bands, were also prevalent \cite{sr13,sr14,sr15,sr16,sr17}.

The late 1980s to 1990s saw enhancements in data storage and processing capabilities, introducing techniques such as band-integrated power Z-scores and ratios after filtering in the 8\-30 Hz range to emphasize seizure dynamics \cite{sr18}, alongside multimodal EEG-video-audio systems for distinguishing epileptic from psychogenic episodes \cite{sr19}. By the 1990s, ESD and ESP transitioned to feature-driven statistical learning, incorporating time-domain, frequency-domain, and wavelet-based features \cite{sr20,sr21,sr22}. Key innovations included nonlinear dynamical tools such as fractal dimensions \cite{sr22}, discrete wavelet decomposition \cite{sr20}, and measures such as Lyapunov exponents and Kolmogorov entropy for prediction \cite{sr21}. The 2000s to 2010s witnessed widespread adoption of machine learning (ML), featuring support vector machines (SVMs) \cite{gonzalez2003support}, artificial neural network (ANN)-based detection for extended EEG recordings \cite{sr23}, singular value decomposition (SVD) for time-frequency features \cite{sr24}, and comparative pipelines using multilayer perceptron neural networks (MLPNNs) and logistic regression \cite{sr25}, as well as hybrid neuro-fuzzy systems and genetic algorithm-optimized classifiers \cite{sr27,sr28,sr29}.

\begin{figure*}[!ht]
 \centering
\includegraphics[width=\linewidth]{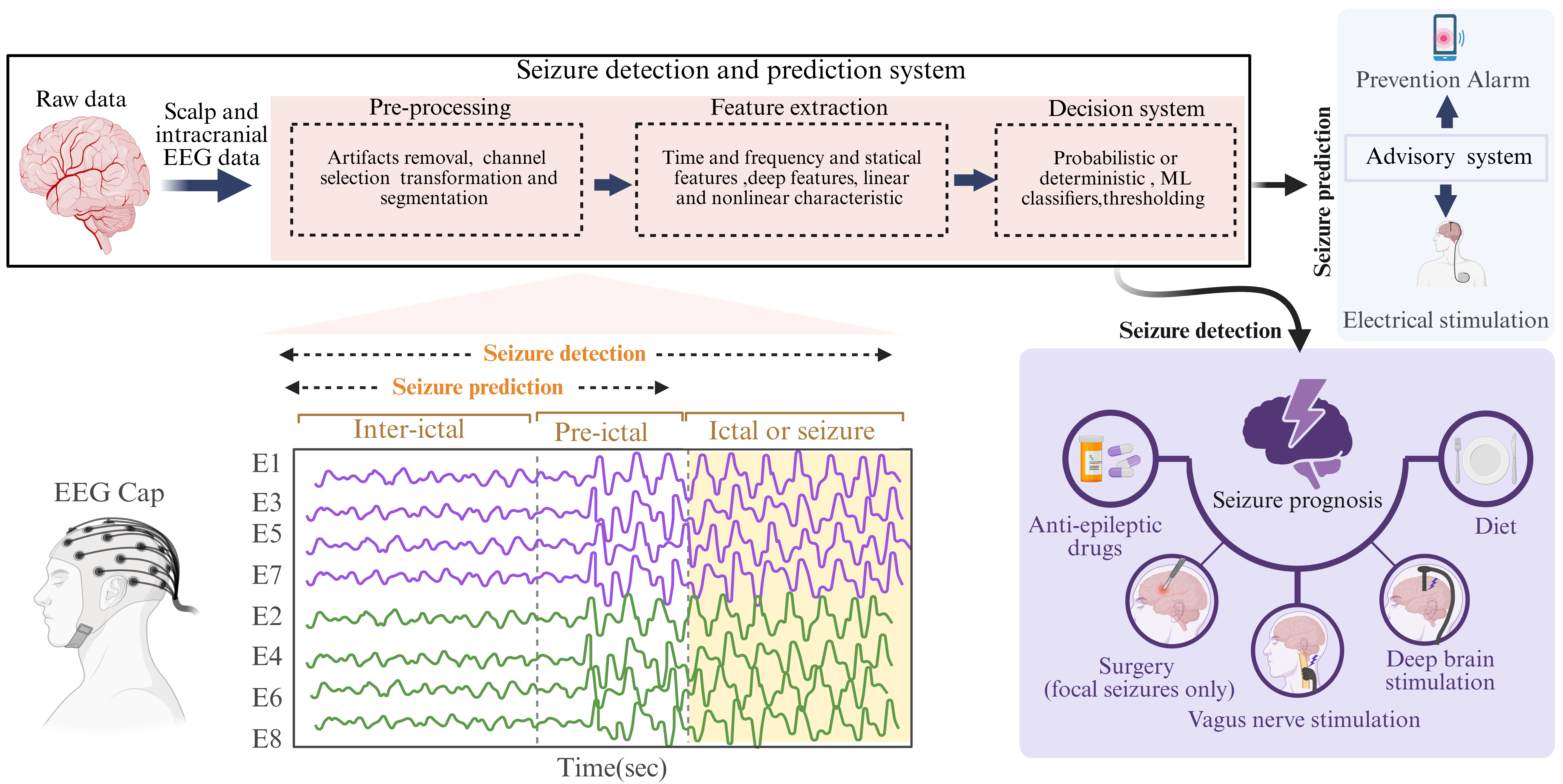}
    \centering
   \caption{ A typical epileptic seizure detection and prediction pipeline: EEG is recorded and preprocessed, key pre-ictal/ictal biomarkers are extracted, and relevant features are fed into a decision module that identifies seizure-related changes and triggers patient alerts. The typical pipeline presents a feature engineering and decision-making system with diagnosis and treatment management.}
  \label{fig:3}
 \end{figure*}

From 2010 to 2020, deep learning (DL) emerged as the dominant paradigm, evolving from patient-specific recurrent neural network (RNN) frameworks \cite{sr30} to deep belief networks \cite{sr31}, stacked autoencoders \cite{sr32}, and end-to-end models like convolutional neural networks (CNNs), long short-term memory (LSTM) networks, and attention mechanisms for raw EEG processing \cite{sr33,sr34,sr35,sr36,sr37,sr38}. Most recently (2020-2025), advanced learning architectures (ALAs) have integrated hybrid pipelines, attention-based networks, foundation model transformers, large language models, and explainable AI (XAI) to enhance robustness and interpretability in seizure detection and prediction  \cite{sr39,sr40,wang2023}. Attention mechanisms facilitate sophisticated spatial-temporal modelling and multimodal fusion, increasingly incorporating heterogeneous signals from wearables alongside EEG for real-world applications \cite{su2025,gill,bha,multi}. Fig. \ref{fig:3} \cite{BioRender} shows the typical ESD and ESP.

Despite these advancements, EEG's inherent limitations, such as susceptibility to artefacts and limited spatial resolution, necessitate multimodal fusion to bolster clinical decision-making, enable heterogeneous data integration, and advance personalised seizure management. For example, incorporating complementary modalities such as electrocardiography (ECG), electromyography (EMG), electrodermal activity (EDA), and photoplethysmography (PPG) captures autonomic and cardiovascular changes associated with seizures, while advanced neuroimaging techniques including magnetoencephalography (MEG), functional near-infrared spectroscopy (fNIRS), positron emission tomography (PET), and functional magnetic resonance imaging (fMRI) provide enhanced spatial and hemodynamic insights to improve detection sensitivity and prediction horizons \cite{sr44,sr49,sr51,sr53,sr57,sr61,sr64,sr76,sr82}. Furthermore, video monitoring synergises with these signals to enable behavioural correlation and artefact rejection, fostering robust, real-time systems for ESD and  ESP in clinical and ambulatory settings \cite{sr76,sr82}. Advanced multimodal monitoring systems, while prevalent in other healthcare domains, have been underutilized in epilepsy. Prior surveys have primarily addressed multimodal devices for seizure detection \cite{sr41,sr42}, with most literature focusing on single-modality approaches \cite{sr43,sr44,sr45,sr46}. Existing multimodal reviews emphasize physiological signals, often overlooking neuroimaging and video-based modalities  \cite{sr41,sr42} as shown in Table \ref{tab-01}. To the best of our knowledge, this is the first comprehensive survey to systematically review multimodal ESD and ESP, encompassing physiological signals, neuroimaging, and video-based monitoring. The scope of this survey extends to advanced multimodal systems in clinical settings, key challenges in multimodal integration, recommendation frameworks, and future directions for both seizure detection and prediction.

 This survey provides a comprehensive review of advanced multimodal learning for seizure detection and prediction (AMLSDP). The primary objective is to transfer technology from EEG or single-modality approaches to wearable, multimodal systems suitable for clinical monitoring to improve disease management and enhance patient safety. We first summarize the historical eras of ESD and ESP, then transition to multimodal monitoring frameworks.
\subsection{Contribution of the paper}
 This survey is unique because it provides comprehensive answers to questions like why multimodality is useful for ESD and ESP and how the evaluation of the relatively newer modalities and pipelines, like physiological signals, neuroimaging, video-based monitoring, novel fusion techniques, advanced feature engineering, and neural decoding models with edge computing, is helpful for ESD and ESP. It discusses the future research and direction in this area.

\begin{table}[t]
\caption{Comparison with Existing Survey Papers. Legends: \checkmark = Discussed, $\times$ = not Discussed}
\label{tab-01}
\centering
\scriptsize
\setlength{\tabcolsep}{2pt}
\renewcommand{\arraystretch}{1.1}
\resizebox{\columnwidth}{!}{%
\begin{tabular}{>{\raggedright\arraybackslash}p{2.2cm}cccccc}
\hline
\textbf{Work} & \textbf{Detection} & \textbf{Prediction} &
\multicolumn{3}{c}{\textbf{Multimodal}} & \textbf{Future direction} \\
\cline{4-6}
 &  &  & \textbf{Physio. signals} & \textbf{Neuroimaging} & \textbf{Video} &  \\
\hline
Alotaiby et al.~\cite{sr43}  & \checkmark & $\times$  & \checkmark  & $\times$ & \checkmark  & $\times$ \\
Chen et al.~\cite{sr41}      & \checkmark & \checkmark  & \checkmark & $\times$ & \checkmark & \checkmark \\
Shoeibi et al.~\cite{sr42}   & \checkmark & \checkmark & $\times$  & \checkmark & \checkmark & \checkmark \\
Rasheed et al.~\cite{sr44}   & \checkmark & $\times$ & \checkmark & $\times$  & $\times$  & \checkmark \\
Kuhlmann et al.\cite{sr45}  & $\times$  & \checkmark & \checkmark & $\times$  & $\times$  & $\times$ \\
Shoka et al.\cite{sr46} & \checkmark  & $\times$ & \checkmark & $\times$  & $\times$  & \checkmark \\
This work                    & \checkmark  & \checkmark & \checkmark & \checkmark & \checkmark & \checkmark \\
\hline
\end{tabular}%
}
\vspace{-2mm} 
\end{table}

\subsection{ Organization of the paper }
The organization of the paper is as  follows: 
As shown in Fig.~\ref {fig:1}, Section I highlights the evolution of ESD (1950 to 2025) and ESP (1975 to 2025), including dataset phases, key publications, workshops, and AI models. Section II discusses the multimodality for ESD and ESP, while Sections III and IV discuss the challenges of multimodal non-EEG-based seizure-monitoring systems. Section V presents an AMLSDP  system that includes novel fusion strategies, advanced feature engineering ,neural decoding methods, evaluation metrics, and edge/real-time computing. Section VI highlights future directions and research opportunities, and Section VII presents the conclusion.

\section{Multi-Modalities for Epileptic Seizure Detection and Prediction}

\subsection{Psychological Signal-based Multimodal System}
\subsubsection{ECG}
Recently, researchers have paid much attention to EEG with ECG for diagnosing epilepsy and guiding therapeutic interventions \cite{sr44}. ECG calculates heart rate variability, which measures the differences between successive R-wave peaks \cite{sr48}. The variation reflects the dynamic interplay between the autonomic nervous system's sympathetic and parasympathetic nervous systems (ANS). ECG's advantages stem from its precision in capturing cardiac responses. Seizures often involve abnormal neuronal electrical activity, affecting the central nervous system's regulation of autonomic functions and leading to dysautonomia. HRV analysis can detect the subtle changes in the ANS before seizure onset\cite{sr49}.The power spectral analysis of HRV signals, focusing mainly on low- and high-frequency powers, provides valuable information about an upcoming seizure. The different frequency bands identify the seizure pattern. The integration of both EEG features, such as time-domain and frequency-domain features, and ECG features, such as RR Interval (RRI), Time-Domain HRV Metrics, and Frequency-Domain HRV metrics,  with advanced deep and machine learning improves seizure detection (high sensitivity) and prediction (reduces false positive rate),  potentially improving the diagnosis process of epilepsy and therapeutic interventions for patients. Table \ref{tab-02}  presents the recent works of multimodal seizure detection and prediction, including EEG with ECG.

\subsubsection{EDA}

EDA measures skin conductance variations from sweat gland activity \cite{sr49}, optimally measured on palms, soles, or wrists for long-term monitoring \cite{sr49}. Seizures induce sweating, reducing skin resistance \cite{sr50}, with responses varying by affected brain region and laterality \cite{sr51}. Integrating EDA with EEG captures neurological and autonomic signals, leveraging ML and DL for improved accuracy and reliability \cite{sr52}. Multimodal EEG-EDA systems, especially in wearables, enhance real-time ambulatory monitoring by mitigating EEG artifacts and incorporating preictal autonomic shifts, reducing false alarms. Table \ref{tab-02}  details multimodal EDA-EEG studies for seizure detection and prediction.

\subsubsection{EMG}

EMG captures the specific pathophysiological changes in muscle activity that occur during seizures \cite{sr53}. Quantitative EMG measures are effective in distinguishing epileptic muscle activation from normal physiological movements and nonepileptic seizures \cite{sr54}. EMG confirms motor involvement, filtering EEG artifacts like chewing or blinks. Less susceptible to environmental noise or electrode issues than the EEG, EMG suits ambulatory settings. Combining EMG with EEG improves seizure detection and prediction precision \cite{sr55}. For instance, Tan et al. fused EEG, EMG, and ECG using deep learning for spatial-temporal feature extraction, achieving 98\% accuracy \cite{sr41}. This multimodal approach enables reliable real-time monitoring and early seizure prediction for timely interventions \cite{sr57}. Table \ref{tab-02} include recent multimodal studies that included EMG with EEG for seizure detection and prediction.

\subsubsection{PPG}

 PPG is a non-invasive optical technique that measures blood volume changes in peripheral tissues, typically via wrist-worn devices, to derive heart rate variability, and pulse transit time (PTT) \cite{sr57}. During seizures, alterations in the autonomic nervous system cause HR acceleration or irregularity, making PPG a valuable peripheral signal for detection and prediction \cite{sr58}. Fusing EEG with modalities like EDA, EMG, accelerometry, or ECG enhances robustness in wearables, reducing motion artifacts and improving specificity through brain-physiological correlations. For example, Thorir Mar et al. applied a BrainFusionNet model to multiple physiological signals, achieving 96.70\% sensitivity and 1.0 FP/h \cite{sr59,sr60}.Table \ref{tab-02} present the PPG with other modalities for seizure detection.

\subsection{Neuroimaging-Based Multimodal System}
\subsubsection{MEG}

MEG was first recorded by Cohen in 1968 using a copper induction coil and, more recently, using superconducting quantum interference devices (SQUIDs) for noise reduction \cite{sr61,sr62,sr63}. Recent systems use ~300 sensors for whole-head coverage, requiring magnetic shielding due to low-amplitude signals. MEG offers superior spatial resolution compared to EEG, with less distortion \cite{sr64,sr65,sr66}, and can be combined with EEG to localize epileptogenic zones and improve postoperative outcomes via MEG cluster resection. MEG has high temporal and spatial precision, enabling the detection of epileptic activity and the identification of biomarkers for seizure prediction.

\subsubsection{fNIRS}

fNIRS is a non-intrusive neuroimaging approach that frequently monitors various brain activities by identifying hemodynamic responses, such as changes in blood oxygenation \cite{rosas2019prediction}. Recently, fNIRS has demonstrated its use in improving the detection of pre-ictal and ictal oxygenation patterns associated with seizures \cite{rein2025wide}. Additionally, the literature indicates that fNIRS can detect improved cerebral oxygenation preceding common tonic-clonic epileptic seizures, enabling early seizure detection \cite{moseley2014increased}.
Researchers have employed ML and DL for improved detection of epileptic seizures. For example, Rosas-Romero et al. \cite{rosas2019prediction} employed convolutional neural networks (CNNs) to predict seizures using fNIRS, achieving average accuracies of 96.9\% and 100\%. However, seizure detection and prediction can be further improved by analyzing multiple modalities together, particularly fNIRS with EEG \cite{eastmond2022deep}. For instance, Sipal et al.\cite{sirpal2019fnirs} achieved an average accuracy of 98.3\% when experimenting with fNIRS and EEG data using the LSTM algorithm, compared to 97.6\% and 97\% accuracies when applied to EEG and fNIRS data, respectively . Similarly, Damseh et al. \cite{damseh2024multimodal} demonstrated that intertwining fNIRS and EEG signals data with improved spectral features can significantly enhance the classification accuracy of multimodal assessment on transformers compared to individually processing fNIRS and EEG data. The fusion of fNIRS and EEG data leverages the strengths of both approaches to improve results. EEG offers intense temporal resolution of electrical activity, while fNIRS provides spatially useful information on cortical hemodynamics.

\subsubsection{PET}

PET plays an essential role in capturing various brain activities by detecting metabolic changes associated with neural activity. To detect epilepsy, the commonly employed radiotracer is fluorodeoxyglucose (FDG), which measures regional glucose metabolism \cite{hodolic201618f}. FDG-PET localizes seizure foci when MRI is inconclusive \cite{aklamanu2025review}, identifying epileptogenic metabolic alterations \cite{hernandez2025preclinical}. PET offers high sensitivity over MRI \cite{aklamanu2025review}. Translocator protein PET (TSPO-PET) excels in drug-resistant epilepsy localization, outperforming MRI and FDG-PET \cite{wang2025refining}. Multimodal integration  boosts diagnostic performance \cite{wang2025refining}. Moreover, to further improve seizure prediction and identification, integrating PET with other image modalities, such as MRI, MEG, or fMRI, can provide more detailed insights into brain activity and help better understand seizures. For example, Wang et al. \cite{wang2025refining} reported that combining MRI with TSPO-PET further enhances diagnostic confidence and detection rate. Similarly, Wang et al. \cite{wang2025artificial, zhu2021multimodal} reported that developing a multimodal model that processes multiple modalities, such as fMRI, CT, and EEG, can increase model accuracy and help realize personalized seizure epilepsy treatment. 
\subsubsection{fMRI}

fMRI plays a crucial role in mapping brain activity by measuring blood oxygen level-dependent (BOLD) signals, which serve as indirect indicators of neuronal activation \cite{sr75}. In epilepsy, fMRI identifies seizure onset zones and networks for diagnosis and surgery \cite{sr76}. High spatial resolution enables precise localization \cite{sr77}. Integration with EEG, MEG, or PPG improves the precision of multimodal assessment \cite{sr77,sr78}.

\subsection{Video Monitoring system}

Video monitoring remains a core component in the clinical evaluation of seizures, offering a reliable method for retrospective assessment of seizure-related events. Typically non-invasive, video detection can be enhanced by affixing infrared markers to specific motion-relevant anatomical sites such as joints or limb \cite{sr79,sr80,sr81}. Challenges included blanket occlusion, which was addressed by using contours or colored sleepwear \cite{sr82}. Patients must remain in view, which is easier at night. Thermal imaging can detect through coverings but has lower resolution and higher cost \cite{sr83,sr84}. Privacy issues arise from data capture and transmission.

Several international research teams, including those based in Belgium, the United States, Germany, Portugal, and Italy, are actively developing automated video-based seizure detection systems . One such example is the SAMi Alert device from a U.S.-based company, explicitly designed for convulsive seizure monitoring. This system requires an Apple device, such as an iPhone or iPad, for video display and alarm notifications.

\begin{table*}[htbp]
\caption{Comprehensive Overview of Multimodal Seizure Detection and Prediction}
\label{tab-02}
\centering
\scriptsize
\renewcommand{\arraystretch}{1.9}
\setlength{\tabcolsep}{2pt}
\resizebox{\textwidth}{!}{ 
\begin{tabular}{p{2.1cm}p{1.2cm}p{2.5cm}p{2.4cm}p{1.6cm}p{2.2cm}p{2.0cm}p{1.8cm}p{1.8cm}}
\hline
\textbf{Reference} & \textbf{Task} & \textbf{Modality} & \textbf{Dataset} & \textbf{Seizure Type} & \textbf{Method} & \textbf{Key Performance} & \textbf{Study Design} & \textbf{Advantage} \\
\hline
Chen et al.\cite{sr41} & Seizure detection & EEG + EMG + ACC + EDA + HR & CHB-MIT + Temple + Clinical & Focal, GTCS & Feature fusion + SVM & SENS: 96.8\%, FPR: 0.18/h & Retrospective multi-center & Comprehensive coverage \\
Wu et al.\cite{sr57} & Seizure detection & EEG + PPG + ACC + EDA & Clinical (20 pts) & Tonic-Clonic, Focal & CNN-LSTM fusion & ACC: 98.2\%, SENS: 97.5\% & Prospective trial & Real-time wearable \\
Ingolfsson et al.\cite{sr60} & Seizure detection & EEG + PPG + ACC & Bonn + Clinical & Focal, Absence & BrainFuseNet & AUC: 0.98, SENS: 95.3\% & Cross-dataset & Edge-deployable \\
Sirpal et al.\cite{sr19} & Seizure detection & EEG + fNIRS & Clinical (8 pts) & Focal, Generalized & Signal correlation & SENS: 100\%, FPR: 0.2/h & Intraoperative & Improved localization \\
Damseh et al.\cite{damseh2024multimodal} & Seizure prediction & EEG + fNIRS & Clinical pediatric & Focal Onset & Vision Transformer & ACC: 97.1\% & Cross-subject & Spatio-temporal capture \\
Theodore et al.\cite{sr22} & Seizure detection & EEG + Video & Long-term VEEG & All types & Fractal analysis & SENS: 92\%, SPEC: 95\% & Preliminary clinical & Video-EEG sync \\
Yang et al.\cite{sr37} & Seizure detection & EEG + Video + ACC & Temple Hospital & Tonic-Clonic, Myoclonic & Multi-view U-Net & F1: 0.96 & Multi-center & Motion artifact robust \\
Boyne et al.\cite{sr84} & Seizure detection & Video + EEG & Clinical (50 seizures) & Tonic-Clonic & 3D CNN & SENS: 98.5\%, SPEC: 99.2\% & Retrospective VEEG & Contactless \\
Hu et al.\cite{sr85} & Seizure detection & Video + EEG & CHB-MIT Video & Focal, Generalized & STMemAE & AUC: 0.99 & Unsupervised & Instance-level detection \\
Banik et al.\cite{sr53} & Seizure detection & EEG + EDA & Emotional tasks & Emotional triggers & EEG-EDA + XAI & ACC: 94\% & Experimental & Explainable link \\
Van de Vel et al.\cite{sr59} & Seizure detection & EEG + ACC + EMG + Audio & Clinical (15 pts) & Tonic-Clonic, Myoclonic & Multi-sensor fusion & SENS: 99\%, FPR: 0.1/h & Home monitoring & SUDEP prevention \\
Yu et al.\cite{sr75} & Seizure detection & EEG + ACC + EDA + HR & Clinical (107 pts) & All types & AI-enhanced fusion & SENS: 95.8\%, FPR: 0.22/h & Large-scale & Wearable scalability \\
Beniczky et al.\cite{sr54} & Seizure detection & EMG + ACC & Clinical (42 pts) & Tonic-Clonic & Wearable EMG & SENS: 98.5\%, FPR: 0.15/h & Prospective & Real-time tonic-clonic \\
Baumgartner et al. \cite{sr55} & Seizure detection & sEMG + ACC & Clinical (89 seizures) & Motor seizures & sEMG analysis & ACC: 92\% & Retrospective & Motor pattern recognition \\
Joshi et al.\cite{sr48} & Seizure detection & HRV + RESP + ACC & Neonatal ICU & Neonatal & ML feature fusion & SENS: 94\%, SPEC: 89\% & Prospective NICU & Non-invasive neonatal \\
Perez-Sanchez et al. \cite{sr49} & Seizure prediction & ECG + HRV & Clinical (18 pts) & Focal & Wavelet + ML & SENS: 91\%, FPR: 0.2/h & Cross-patient & ECG-based prediction \\
Ortega et al.\ \cite{sr50} & Seizure detection & EDA & Meta-analysis (12 studies) & All types & EDA response & SENS: 85\% & Systematic review & Autonomic marker \\
Halimeh et al.\ \cite{sr51} & Seizure detection & EDA + HR + HRV & Clinical (25 pts) & All types & Wearable analysis & ACC: 88\% & Longitudinal & Medication monitoring \\
Horinouchi et al.\cite{sr52} & Seizure detection & EDA & Clinical (30 pts) & Epilepsy & Baseline comparison & EDA reduction: 65\% & Case control & Diagnostic biomarker \\
Rémi et al.\cite{sr54} & Seizure detection & Video + Motion & Clinical (45 pts) & Focal automatisms & 3D motion analysis & ACC: 95\% & Retrospective & Automatism quantification \\
Loesch-Biffar et al.\cite{sr82} & Seizure detections & Video + 3D Motion & Clinical (60 seizures) & Ictal automatisms & 3D visualization & Detection: 97\% & Prospective & Clinical visualization \\
Faust et al.\cite{faust2025detecting} & Seizure detection & ACC + PPG + EDA & Clinical (120 pts) & All types & ML comparison & SENS: 96.2\%, FPR: 0.12/h & Multi-device & Device-agnostic \\
Yu et al.\cite{yu2023artificial} & Seizure detection & ACC + EDA + HR & Clinical (107 pts) & All types & Wearable AI & SENS: 92.5\%, FPR: 0.3/h & Large-scale & Standalone wearable \\
Van de Vel et al.\cite{sr59} \textsuperscript{*} & Seizure detection & ACC + EMG + Audio & Clinical (15 pts) & Tonic-Clonic & Sensor fusion & SENS: 97\%, FPR: 0.15/h & Home monitoring & Non-EEG SUDEP \\
Karasmanoglou et al.\cite{karasmanoglou2025semi } & Seizure prediction & EEG + ECG & Clinical pediatric (7 pts) & Focal & Semi-supervised anomaly detection & AUC: 95\% & Retrospective & Wearable, low false alarms \\
Xiong et al.\cite{xiong2023forecasting} & Seizure prediction & Self-reported + HR & Clinical (13 pts) & All types & Cycle extraction + regression & AUC: 0.77 & Prospective pilot & Robust, generalizable \\
Hadipour et al.\cite{hadipour2021predicting} & Seizure prediction & IMU (smart glasses) & Simulated & Epileptic & LSTM & ACC: high & Experimental & Non-invasive \\
Boyne et al.\cite{sr84}\textsuperscript{*} & Seizure detection & Video (3D CNN) & Clinical (50 seizures) & Tonic-Clonic & 3D CNN & SENS: 97.8\%, SPEC: 98.5\% & Retrospective & Pure video detection \\
\hline
\end{tabular}
}
\vspace{2pt}
\scriptsize \emph{\textsuperscript{*}mark shows non-EEG-based multimodalities.}
\end{table*}

\section{Challenges of multimodal epileptic seizure detection and prediction}

\subsection{Data Scarcity and Heterogeneity}

Data scarcity and heterogeneity pose significant challenges in multimodal epileptic seizure detection and prediction . Public datasets like CHB-MIT, Bonn, and EPILEPSIAE are limited in size, duration, and diversity, often acquired in controlled settings that cannot capture real-world variability \cite{zhang2024review}. This scarcity is exacerbated in multimodal studies, where synchronised acquisition of EEG with other biosignals (e.g., ECG, EMG, or fNIRS) is technically demanding and resource-intensive. The resulting imbalance and underrepresentation of seizure types, particularly focal and non-convulsive events, hinder robust model generalization. Moreover, inter-patient and intra-patient variability in seizure morphology, sensor placement, and environmental noise complicate the development of universal predictive frameworks \cite{kerr2024present,sr59}.

Addressing data scarcity also necessitates tackling heterogeneity in acquisition protocols, feature representations, and data quality across studies and devices. Non-EEG modalities introduce additional challenges due to differences in sampling rates, signal-to-noise ratios, and physiological latency, making multimodal fusion and model alignment difficult.
Heterogeneity in protocols, features, and quality across studies exacerbates domain shifts, degrading performance on unseen data \cite{edelman2024non}. Mitigation strategies include few-shot learning for generalisation with minimal data \cite{li2024survey}, and GAN-based or self-supervised data augmentation to enhance datasets while maintaining fidelity \cite{zhang2024review}. However, the lack of large-scale, standardized multimodal benchmarks limits reproducible research and clinical translation \cite{kerr2024present}.

\subsection{Integration of Non-Invasive Biomarkers}

Integrating non-invasive biomarkers (EEG, ECG, EDA, EMG, PPG, fNIRS, MEG, fMRI) into unified frameworks is challenging due to synchronization, calibration, and fusion issues. Signals differ in temporal dynamics, sampling rates, and noise, complicating alignment \cite{zhang2024review,edelman2024non}. Non-EEG modalities are prone to motion artifacts and environmental variations, potentially obscuring patterns. The absence of standardized protocols and benchmarks hinders reproducibility \cite{kerr2024present}.

From an analytical standpoint, heterogeneous signal characteristics demand advanced multimodal fusion strategies that can jointly exploit complementary features while mitigating redundancy and noise. Early-fusion techniques (feature-level concatenation) risk information dilution due to differing scales and noise distributions, whereas late fusion (decision-level integration) may lose fine temporal dependencies between modalities \cite{edelman2024non}. Intermediate or hybrid fusion methods, often employing deep learning architectures such as convolutional, recurrent or attention-based networks, have shown promise in learning cross-modal representations \cite{zhang2024review}. Nonetheless, these models remain data-hungry and prone to overfitting, especially when trained on small or modality-imbalanced datasets. Recent approaches leveraging transfer learning and few-shot paradigms attempt to address the limited availability of multimodal samples by transferring knowledge from unimodal tasks or synthetic data \cite{li2024survey}. Despite these advances, a robust physiological interpretation of integrated biomarkers is still lacking, hindering clinical trust and regulatory approval. Establishing standardised pipelines for multimodal data acquisition, alignment, and interpretability, potentially through open-access frameworks and IoMT-enabled monitoring systems, will be essential for the translational success of non-invasive biomarker integration \cite{kerr2024present}.

\subsection	{Real-Time Processing Latency}
Achieving real-time seizure detection and prediction remains a major technical challenge in multimodal systems due to constraints in computational efficiency, data transmission, and energy consumption. Continuous monitoring of high-dimensional signals, such as EEG, ECG, EMG, and fNIRS, demands rapid acquisition, preprocessing, and classification pipelines capable of handling large data streams without compromising temporal resolution \cite{zhang2024review, kerr2024present}. The intrinsic latency between data acquisition, feature extraction, and decision-making can delay seizure alerts, reducing their clinical relevance for early intervention. In wearable and Internet of Medical Things (IoMT) architectures, latency is further aggravated by wireless data transfer, sensor synchronization, and cloud-based computation \cite{edelman2024non}. While deep neural networks have substantially improved prediction accuracy, their computational complexity often makes real-time deployment impractical on edge or embedded devices with limited processing power and memory \cite{li2024survey}.

Addressing these latency constraints requires a trade-off between computational depth and response speed. Lightweight architectures, such as convolutional and recurrent neural networks optimized through model pruning, quantization, or knowledge distillation, have shown promise for on-device seizure forecasting with reduced inference times \cite{zhang2024review}. Edge computing frameworks are increasingly adopted to perform localized preprocessing and decision-making near the data source, minimizing transmission delay and improving privacy compliance \cite{edelman2024non}. However, maintaining synchronization across multimodal sensors remains nontrivial, as disparate sampling rates and asynchronous data streams can introduce temporal misalignment, thereby affecting classification accuracy. Ultimately, the design of future multimodal systems must balance predictive accuracy with system responsiveness to ensure both timely intervention and user comfort in clinical and ambulatory environments.

\subsection	{Model Generalization and Inter-Patient Variability}
Developing seizure detection and prediction systems that generalize across patients, datasets, and recording environments remains one of the most persistent challenges in multimodal epilepsy research. Physiological signals such as EEG, ECG, and EMG exhibit strong inter- and intra-subject variability, influenced by age, seizure type, medication status, sleep cycles, and comorbidities \cite{kerr2024present}. Deep learning models often capture subject-specific idiosyncrasies rather than universal biomarkers, resulting in significant performance degradation when tested on unseen individuals or different recording systems \cite{zhang2024review}. This lack of generalization has hindered large-scale clinical deployment, as retraining models for each new patient is computationally expensive and clinically impractical. Moreover, inconsistencies in electrode placement, sensor calibration, and recording duration further exacerbate model bias and limit transferability across institutions and hardware platforms \cite{edelman2024non, halimeh2023explainable, faust2025detecting, yu2023artificial}.

To overcome these limitations, researchers have increasingly explored domain adaptation, meta-learning, and self-supervised learning  to enhance cross-subject generalization and robustness. Domain adaptation approaches, such as adversarial and feature-invariant training, enable models to learn patient-independent representations that mitigate signal heterogeneity across recording sessions and acquisition setups \cite{tazaki2025eeg}. Similarly, meta-learning and few-shot paradigms facilitate rapid personalization of pretrained models to new patients with minimal calibration data, allowing models to adapt to inter-patient variability without exhaustive retraining \cite{li2024survey}. SSL has also emerged as a robust framework for leveraging vast unlabeled biosignal datasets to learn transferable representations, significantly improving downstream seizure detection performance under limited supervision \cite{das2022improving}. Furthermore, recent studies demonstrate that hybrid CNN–BiLSTM or transformer-based architectures combined with adversarial training can enhance the generalization of multimodal EEG systems across diverse patient populations \cite{zhang2024efficient,sigsgaard2023improving}. Despite these advances, achieving clinically reliable cross-patient generalization continues to require standardized acquisition protocols, multimodal benchmark datasets, and systematic validation across heterogeneous clinical cohorts.

\subsection{Artificial Intelligence Interpretability and Clinical Trust}
Despite remarkable progress in the accuracy of deep learning–based seizure detection and prediction systems, AI interpretability and clinical trust remain significant barriers to their adoption in real-world healthcare. Most multimodal seizure prediction frameworks employ complex neural architectures that act as opaque "black boxes," providing limited insight into how multimodal features such as EEG, ECG, or EDA contribute to model decisions \cite{kerr2024present, holzinger2019causability}. This lack of transparency poses significant risks in clinical decision-making, where explainability is essential for understanding model outputs and ensuring accountability. In epilepsy management, clinicians demand not only accurate predictions but also physiological plausibility and clear reasoning that relates computational features to known seizure biomarkers and pathophysiology \cite{ribeiro2016should},without transparent interpretability mechanisms, even high-performing models struggle to achieve clinical credibility or regulatory validation.

Recent advances in XAI are helping bridge this gap by making AI-driven neurodiagnostic systems more transparent and trustworthy. Visualization-based methods, including saliency maps and layer-wise relevance propagation, enable clinicians to observe which temporal–spatial regions of EEG or related biosignals influence a model’s decision, improving diagnostic confidence \cite{cui2023towards, tjoa2020survey}. Model-agnostic explanation frameworks such as SHAP (SHapley Additive exPlanations) and  LIME (Local Interpretable Model-agnostic Explanations)  have been applied to quantify feature importance across modalities, highlighting which physiological signals contribute most to predicted seizure onset \cite{ribeiro2016should}. Moreover, recent work emphasizes causability, or the ability to provide reasoning that is both interpretable to clinicians and causally consistent with medical knowledge \cite{holzinger2019causability}. Broadly, explainable AI enhances transparency, model safety, and human oversight by integrating physicians into the interpretation loop. However, challenges remain in standardizing interpretability metrics, ensuring robustness of explanations under noisy multimodal inputs, and balancing computational efficiency with explainability in real-time systems \cite{sadeghi2024review}. Achieving clinically trustworthy multimodal seizure prediction thus requires a unified XAI framework combining transparency, causality, and user-centric validation.

\section{Challenges of non-EEG multimodal-based epileptic seizure detection and prediction}

Non-EEG-based seizure detection and prediction systems employ non-invasive, wearable signals such as EMG, ACM, ECG, EDA, and PPG as alternatives to traditional EEG. These methods capture physiological responses to seizures and allow real-time alerts to help prevent or reduce the risk of SUDEP. However, without EEG or neural signals, the performance of the real-time seizure detection and prediction system is affected. The key challenges are as follows:
\subsection{Data Scarcity of Non-EEG multimodality }
A significant challenge in advancing non-EEG-based seizure detection and prediction is the scarcity of publicly available datasets that include modalities such as EMG, PPG, ECG, and other physiological signals. Existing datasets, such as the 'Seizure Gauge Wearable' and the Open Seizure Database, have limitations in their coverage of seizure types, collection conditions, and clinical semiology \cite{sr87}. Resource-intensive collection via specialized equipment and expert annotation in signal processing and epileptology exacerbate issues \cite{sr88,sr89}. Collaboration among experts and standardized protocols are essential to leverage wearable data for epilepsy monitoring.

\subsection{Data Acquisition of Non-Electrophysiological signals}

The acquisition of non-electrophysiological signals for seizure detection poses significant challenges due to inter- and intra-patient variability and incomplete data. In contrast to EEG, which directly measures neural activity and provides comprehensive seizure coverage, non-EEG modalities serve as indirect indicators, such as heart rate variability, autonomic responses, and movement \cite{yu2023artificial}. Non-electrophysiological signals often fail to capture all seizure types, particularly non-convulsive or focal seizures that lack prominent motor symptoms. They usually miss non-convulsive or focal seizures without motor symptoms, with wearables excelling at detecting tonic-clonic seizures but lacking sensitivity to spikes. Non-seizure data predominance degrades models, while real-world collection is costly. Confounders like daily activities or arousal affect accelerometer or PPG signals.

Standardized devices (e.g., Empatica E4, Apple Watch) with synchronized sensors ensure consistent rates (e.g., 64 Hz). Protocols like Empatica Embrace2 enable 24/7 monitoring with automated tonic-clonic triggers \cite{sr90,sr91,sr92}. Nonetheless, standardized protocols are needed for generalizability across populations.

\subsection{Multimodal Data Integration}

Integrating heterogeneous non-EEG modalities poses challenges due to varying channel configurations, sampling rates, and resolutions. Synchronization is critical for ictal physiological change extraction. For instance, ECG (100 to 250 Hz) contrasts with higher PPG/EMG rates, causing feature imbalances. Solutions involve standardized rates and robust fusion frameworks \cite{sr92,sr93} to enhance wearable epilepsy monitoring applicability.

\subsection{Morphological Features of Seizures from Non-Electrophysiological Signals}

Identifying efficient features from non-EEG modalities for seizure detection and prediction remains a significant challenge, particularly in distinguishing ictal and preictal patterns \cite{sr24,gonzalez2003support,sr37}. Non-EEG relies on peripheral indicators: ECG (RR interval (the time between two consecutive R-waves), frequency ratios), EMG, ACM, EDA, PPG \cite{sr42,sr43,sr50,sr51}. Inter-/intra-patient variability complicates reliability. Robust fusion and feature engineering are required for ambulatory clinical use.

\subsection{Non-EEG Artificial Intelligence Model Challenges}

AI, including machine and deep learning, shows promise in non-EEG seizure detection but faces generalizability, reliability, and translation issues.

\subsubsection{Model Architecture and Explainability}

Advanced deep learning, including transformer-based models, hybrid deep learning, and generative AI models, faces notable challenges due to their complex architectures that leverage non-EEG-based multimodality. The structure involves a high number of parameters and layers. This complexity demands significant computational power, which can limit the models’ usability in
clinical settings due to resource constraints. Additionally, the
black-box nature of deep learning models limited insight into how multimodal features such as  ECG, PPG, or EDA contribute to model decisions \cite{kerr2024present, holzinger2019causability}. The opacity of decision-making processes is
problematic in clinical environments, where explainability is
crucial \cite{sr94}. Clinicians require transparency to justify AI-driven decisions \cite{sr95,sr96}. To overcome these challenges, different approaches have been adopted, such as pruning (removing less important model weights) \cite{sr97}, quantization (reducing model precision to 4 or 8 bits) \cite{sr98}, and knowledge distillation (transferring insights from a large neural network, such as a Transformer, to a smaller model) \cite{sr99}. Lightweight models (with fewer parameters) and edge computing via federated learning (distributed training across local devices to reduce reliance on the cloud) are also applied \cite{sr100}. For explainability, SHAP quantifies how much each feature contributes (e.g., ECG HRV low-frequency power contributes +40\%, ACC variance +30\%) \cite{sr101},LIME approximates local model behavior using interpretable rules (e.g., high heart rate variability combined with an electrodermal activity spike indicates a seizure) \cite{sr101}, and Grad-CAM (Gradient-weighted Class Activation Mapping) visualizes which time windows or sensors in convolutional networks are most important (for example, highlighting a segment of the ECG signal five seconds before an event) \cite{sr102} and the attention map \cite{sr103}. These approaches need to apply to non-EEG-based seizure detection and prediction.

\subsection{Deployment and Clinical Applicability}

Deploying AI in non-EEG devices balances computation, battery life, and adherence \cite{sr104}. Real-time edge processing, sensor displacement, and irritation affect data integrity. Phase-4 trials in heterogeneous populations are lacking, impeding approval \cite{sr105}. Models must align with patient diaries and reduce SUDEP risks. Human-in-the-loop systems and collaborative frameworks are vital for clinical integration.

\section{Advanced Multimodal Monitoring Systems for Epileptic Seizure Detection and Prediction }

\begin{figure*}[!ht]
\centering
\includegraphics[width=\linewidth]{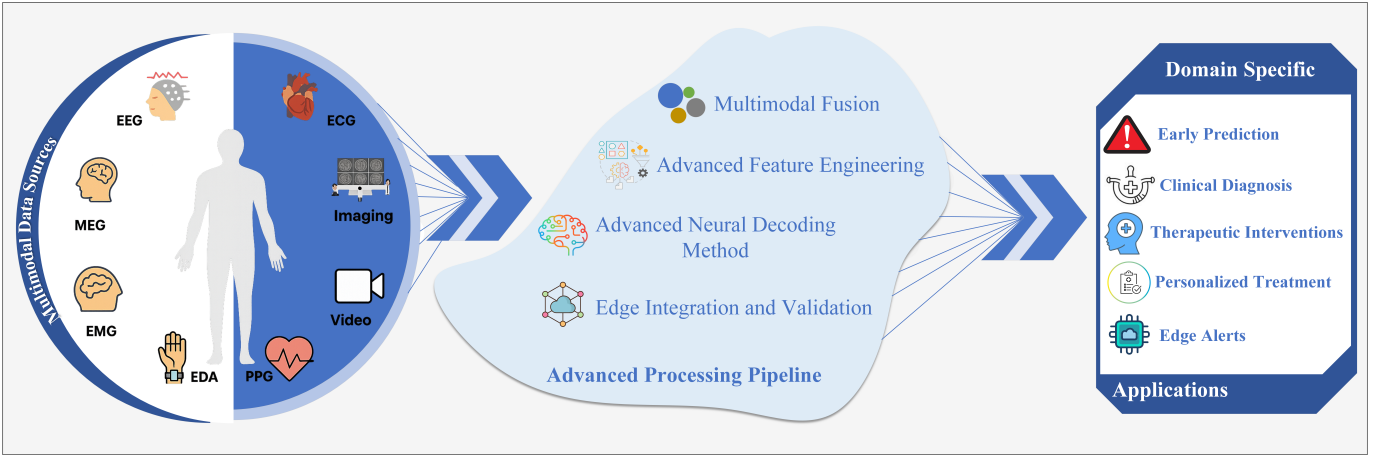}
\caption{Advanced multimodal system for seizure detection and prediction integrating physiological, imaging, and video data. The pipeline includes multimodal fusion, feature engineering, neural decoding, and edge-based validation, enabling early prediction, clinical diagnosis, personalized treatment, and real-time alerts.}
\label{fig-04}
\end{figure*}

This section outlines the primary components of the advanced multimodal seizure detection and prediction system.
These components collectively support ESD and ESP by incorporating multimodal data sources, advanced feature engineering with fusion strategies, advanced neural decoding methods, and post-processing techniques. The multimodal data source is already discussed in Section 2. 
These technologies enable real-time SDP. The AMSDP system aims to improve sensitivity and reduce false alarms for various seizure types, including focal seizures, generalized tonic-clonic seizures , and non-convulsive seizures. Additionally, the system is shown to increase patient compliance and support integration into clinical practice, as shown in Figure \ref{fig-04}.

\subsection{Advanced Fusion Strategies for Multimodal Epileptic Seizure Detection and Prediction System}
\begin{figure}[!ht]
 \centering
\includegraphics[width=3.4in, height=5in]{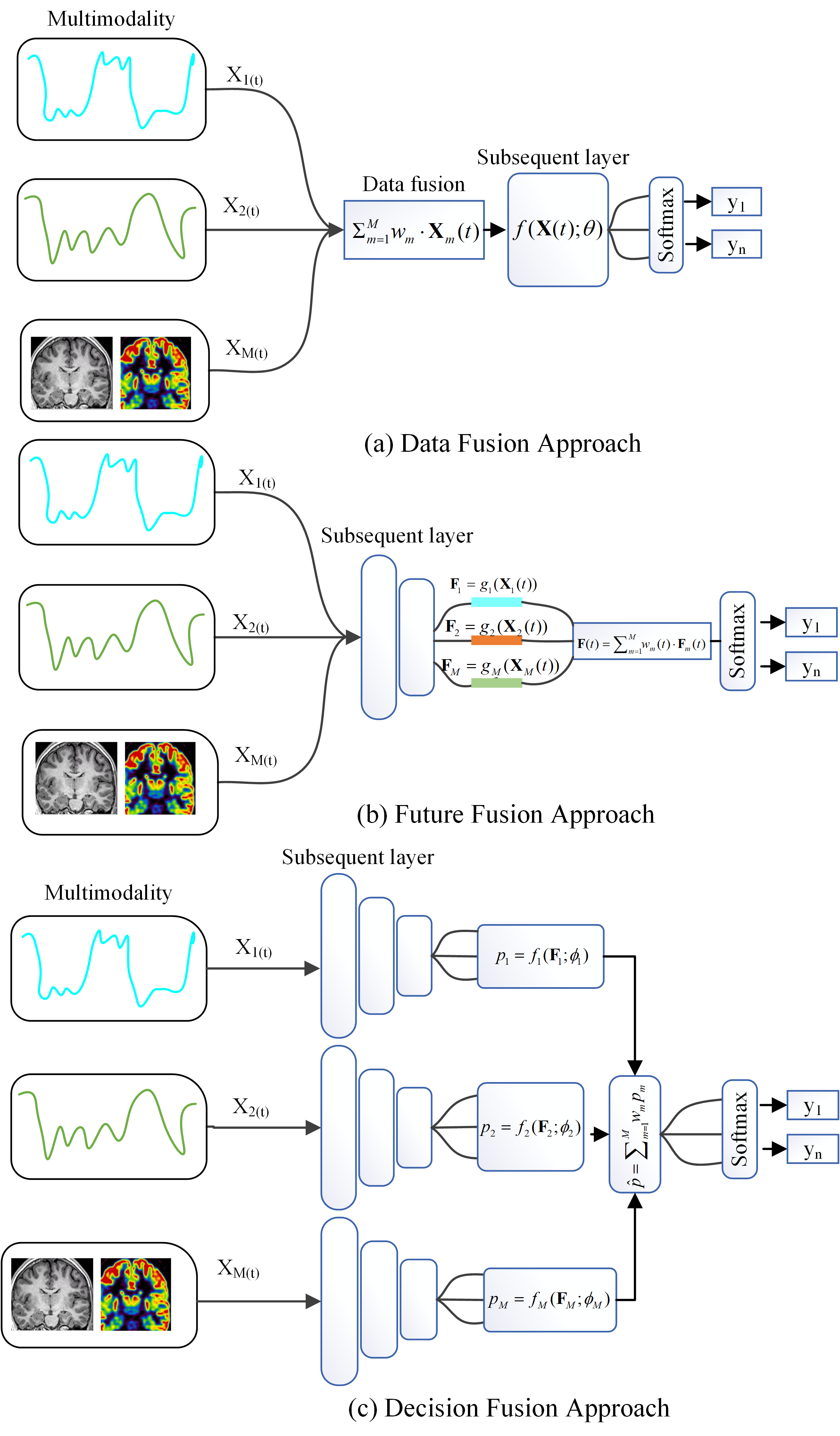}
    \centering
  \caption{Advanced multimodal fusion strategies include (a) early fusion (combining data before processing), (b) intermediate fusion (combining features at an intermediate layer), and (c) late fusion (combining predictions from separate models). }
  \label{fig:fusion}
 \end{figure}

Multimodal data fusion is essential for ESD and ESP, integrating diverse sources to boost accuracy, reduce false positives, and enable proactive intervention. Multimodal fusion approaches are critical for efficient ESD and ESP. There are three levels of fusion: data, feature, and decision-level, using techniques such as data concatenation, weighted integration, logical operations, and deep learning to achieve cross-modal representation. Each modality provides unique data and characteristics about seizures. All data streams are resampled to a standard rate (e.g., 100 Hz) and synchronized using timestamp alignment and cross-correlation. This corrects for sampling differences and sensor clock drift, ensuring temporal coherence across modalities as shown in Figure \ref{fig:fusion}.
\subsubsection{Data Level Fusion}
Data level fusion integrates raw multimodal data, such as physiological signals (e.g., EEG, ECG), imaging, and video streams, or their weighted representations from \(M\) independent modalities, each denoted as \(\mathbf{X}_m(t)\) at time \(t\) \cite{sr106}. This method enhances ESD and ESP sensitivity by leveraging complementary information across modalities, enabling early detection of subtle patterns that are undetectable in any single modality. Recent advances have demonstrated the efficacy of data-level fusion for improving seizure analysis, often incorporating transformations such as wavelet decompositions before concatenation \cite{sr106}. For a temporal window of length \(T\), the fused multimodal data \(\mathbf{X}(t)\) is formulated as:
\begin{equation}
\mathbf{X}(t) = \begin{bmatrix}
\mathbf{X}_{1}(t) \\
\mathbf{X}_{2}(t) \\
\vdots \\
\mathbf{X}_{M}(t)
\end{bmatrix} \quad \text{or} \quad \mathbf{X}(t) = \sum_{m=1}^M w_m \cdot \mathbf{X}_m(t),
\end{equation}
where \(\mathbf{X}(t)\) represents the fused multimodal data vector at time \(t\); \(\mathbf{X}_m(t)\) denotes the raw data from the \(m\)-th modality at time \(t\); \(M\) is the total number of modalities; and \(w_m\) are the attention-derived weights that dynamically emphasize salient contributions from each modality. The resulting fused input \(\mathbf{X}(t)\) encapsulates interactions, such as EEG spikes (\(\mathbf{X}_{1}(t)\)) synchronized with heart rate variability fluctuations in ECG (\(\mathbf{X}_{2}(t)\)), critical for preictal forecasting. In seizure prediction workflows, \(\mathbf{X}(t)\) is segmented into sliding windows and fed into a predictive model \(f(\mathbf{X}(t); \theta)\), yielding the seizure probability \(p(\text{seizure} \mid t)\).
\subsubsection{Feature Level (Intermediate) Fusion}
Feature-level fusion is crucial for ESD and ESP, integrating post-extracted modality-specific features (e.g., EEG spectral power, ECG heart rate variability, video motion artifacts) to capture inter-modal correlations, enhance generalization, mitigate noise, and improve accuracy. In feature-level fusion, features are extracted from each modality and concatenated into a unified feature vector or embedding space \cite{sr57,srij}. Specifically, extracted features \(\mathbf{F}_m \in \mathbb{R}^{D_m}\) from \(M\) modalities are aggregated into a joint vector \(\mathbf{F} \in \mathbb{R}^D\), where \(D = \sum D_m\). This aggregation is typically achieved through weighted concatenation or linear combination. The process of feature-level fusion can be mathematically represented as follows:
\begin{equation}
\mathbf{F} = \left[ \mathbf{F}_{1}; \mathbf{F}_{2}; \dots; \mathbf{F}_{M} \right] \quad \text{or} \quad \mathbf{F} = \sum_{m=1}^M w_m \cdot \mathbf{F}_m
\end{equation}
where \(\mathbf{F}\) is the fused feature vector; \(\mathbf{F}_m\) represents the feature vector extracted from the \(m\)-th modality; \(M\) is the total number of modalities; \(D_m\) is the dimensionality of features from the \(m\)-th modality; \(D\) is the total dimensionality of the fused features; and \(w_m \geq 0\) denotes the modality weights, which can be determined using softmax attention, where \(w_m = \frac{\exp(a_m)}{\sum \exp(a_k)}\) and \(a_m\) represents the attention score for the \(m\)-th modality. In the case of non-linear fusion, such as through neural network layers, the fused representation is given by \(\mathbf{F} = g(\{\mathbf{F}_m\}_{m=1}^M; \theta)\), where \(g\) denotes a fusion network (e.g., a multilayer perceptron) and \(\theta\) are its parameters. The fused representation \(\mathbf{F}\) is subsequently provided as input to a classifier \(f(\mathbf{F}; \phi)\), which outputs the probability of a seizure, denoted as \(p(\text{ictal} | \mathbf{F})\).
\subsubsection{Decision-Level (Late) Fusion}
Decision-level fusion is vital for ESD and ESP, as it aggregates independent modality-specific decisions \cite{sr108}. In this approach, independent classifiers or models analyze features extracted from each data modality separately. Each model generates outputs specific to its modality, including probability scores, class labels, or confidence estimates. The system then aggregates these outputs to form a unified decision. Decision-level fusion integrates outputs from \(M\) modality-specific models, where each output is represented as either \(p_m\) seizure probability or \(l_m\) class label. The following equation provides a mathematical representation of this process.
\begin{equation}
\hat{p} = \sum_{m=1}^M w_m p_m \quad \text{(weighted sum)},
\end{equation}
or
\begin{equation}
\hat{l} = \arg\max_k \sum_{m=1}^M \mathbb{I}(l_m = k) \quad \text{(majority voting)},
\end{equation}
where \(\hat{p}\) is the fused seizure probability; \(\hat{l}\) is the fused class label; \(p_m\) is the probability output from the \(m\)-th modality-specific model; \(l_m\) is the class label from the \(m\)-th modality-specific model; \(M\) is the total number of modalities; \(w_m\) are the reliability weights for each modality; \(\mathbb{I}(\cdot)\) is the indicator function that equals 1 if the condition is true and 0 otherwise; and \(k\) represents the possible class indices (e.g., ictal, interictal). The detailed explanation is given in Table \ref{tab-07}.
\subsubsection{Advanced Feature Engineering}
Advanced feature engineering is pivotal for high-performance MSDP. Contemporary methods construct temporally aligned feature sets from EEG, ECG, EMG, EDA, PPG, fNIRS, and video streams, forming a concatenated multimodal feature vector:
\begin{equation}
\mathbf{F}(t) = [\mathbf{F}_{1}(t)^\top, \mathbf{F}_{2}(t)^\top, \mathbf{F}_{3}(t)^\top, \ldots]^\top \in \mathbb{R}^D
\end{equation}
where \(\mathbf{F}(t)\) is the concatenated multimodal feature vector at time \(t\); \(\mathbf{F}_m(t)\) denotes the feature vector from the \(m\)-th modality (e.g., \(\mathbf{F}_1(t)\) for EEG, \(\mathbf{F}_2(t)\) for ECG, \(\mathbf{F}_3(t)\) for video) at time \(t\); \(^\top\) indicates the transpose operation; and \(D\) is the total dimensionality of the fused features. Modality-specific features include spectral power, Hjorth parameters, wavelet energy, HRV indices (RMSSD, LF/HF), entropy, and deep video embeddings (3D-CNN or pose estimation). Cross-modal dependencies are captured via second-order statistics, cross-frequency coupling, phase-amplitude coupling, and Granger causality. Attention-weighted fusion is commonly applied:
\begin{equation}
    \mathbf{F}(t) = \sum_{m=1}^{M} w_m(t) \cdot \mathbf{F}_m(t)
\end{equation}
where \(\mathbf{F}(t)\) is the weighted fused feature vector at time \(t\); \(\mathbf{F}_m(t)\) is the feature vector from the \(m\)-th modality at time \(t\); \(M\) is the total number of modalities; and \(w_m(t)\) are the dynamic attention weights at time \(t\) that emphasize preictal-relevant modalities. These engineered representations significantly outperform unimodal baselines, reducing false alarm rates by up to 40\% and extending prediction horizons >60 min while maintaining sensitivity >85\%.
\subsection{Advanced Neural Decoding Methods}
Following the discussion of multimodal data, feature-level, and decision-level fusion approaches, the subsequent phase focuses on advanced neural decoding methods for seizure detection and prediction. These methods leverage deep learning, transfer learning, and adaptive learning methods for ESD and ESP, as shown in Table \ref{tab-03}.
\subsubsection{Deep Learning Methods}
DL, a specialized subset of ML that uses artificial neural networks, enables end-to-end representation learning and delivers superior performance in areas such as speech recognition and computer vision \cite{sr108}. DL models autonomously derive hierarchical features from multimodal data, achieving state-of-the-art performance in SDP tasks \cite{sr109}. Notably, convolutional neural networks (CNNs) adeptly discern intricate patterns in physiological signals and imaging modalities, thereby reducing inter-patient variability and lowering false-positive rates \cite{ahmad2024attention}. In multimodal systems, CNNs process raw data \(\mathbf{X} \in \mathbb{R}^{C \times T}\) via convolutional layers:
\begin{equation}
Z_{c,t} = \sigma \left( \sum_{m=1}^{C} \sum_{\tau=-k}^{k} W_{c,m,\tau} \, X_{m,t-\tau} + b_c \right)
\end{equation}
where \(Z_{c,t}\) is the output feature map at channel \(c\) and time \(t\); \(\sigma\) is the activation function (e.g., ReLU); \(C\) is the number of input channels; \(W_{c,m,\tau}\) are the weights of the convolution kernel for output channel \(c\), input channel \(m\), and temporal offset \(\tau\); \(X_{m,t-\tau}\) is the input data at channel \(m\) and shifted time \(t-\tau\); \(k\) defines the kernel size (half-width); \(b_c\) is the bias term for channel \(c\); and the summation performs the convolution operation. Followed by a pooling layer:
\begin{equation}
\hat{Z}_{c,t} = \operatorname{pool}_{\mathcal{N}(t)} \{ Z_{c,u} \}_{u \in \mathcal{N}(t)}
\end{equation}
where \(\hat{Z}_{c,t}\) is the pooled output at channel \(c\) and time \(t\); \(\operatorname{pool}\) is the pooling function (e.g., max or average); \(\mathcal{N}(t)\) is the neighborhood around time \(t\) for pooling; and \(Z_{c,u}\) are the values in the feature map within the neighborhood. The explanation of the equation is given in the literature \cite{ahmad2024attention}. At the same time, Long short-term memory (LSTM) networks address temporal dependencies and vanishing gradients with input, forget, and output gates \cite{ahmad2023hybrid}. The fully connected layers classify outputs into ictal, interictal, or preictal states. For instance, hybrid CNN-LSTM models yield superior results for seizure detection \cite{aboyeji2025dcsenets}\cite{ahmad2023hybrid}. Despite the effectiveness of spectral and temporal features extracted by CNNs and LSTMs, challenges remain in focusing on the crucial elements of multimodal representations with long-range dependencies. Attention mechanisms are used to focus on salient input regions \cite{he2024seizurelstm}, with Transformers employing multi-head attention (MHA) to capture global dependencies \cite{li2024anomaly}. The mathematical expressions are as follows:
\begin{equation}
\operatorname{Attn}(\mathbf{Q},\mathbf{K},\mathbf{V}) = \operatorname{softmax} \left( \frac{\mathbf{Q} \mathbf{K}^\top}{\sqrt{d_k}} \right) \mathbf{V}
\end{equation}
where \(\operatorname{Attn}\) is the attention output; \(\mathbf{Q}\), \(\mathbf{K}\), and \(\mathbf{V}\) are the query, key, and value matrices, respectively; \(\operatorname{softmax}\) normalizes the attention scores; \(^\top\) denotes transpose; and \(\sqrt{d_k}\) is the scaling factor with \(d_k\) being the dimension of the keys.
\begin{align}
\text{MHA}(\mathbf{X}) &= \operatorname{Concat}(\text{head}_1, \ldots, \text{head}_H) \, \mathbf{W}^O \\
\text{head}_h &= \operatorname{Attn} \big( \mathbf{X} \mathbf{W}_h^Q, \, \mathbf{X} \mathbf{W}_h^K, \, \mathbf{X} \mathbf{W}_h^V \big)
\end{align}
where \(\text{MHA}\) is the multi-head attention output; \(\mathbf{X}\) is the input matrix; \(\operatorname{Concat}\) concatenates the outputs from multiple heads; \(H\) is the number of attention heads; \(\mathbf{W}^O\) is the output projection matrix; \(\text{head}_h\) is the output from the \(h\)-th head; and \(\mathbf{W}_h^Q\), \(\mathbf{W}_h^K\), \(\mathbf{W}_h^V\) are the projection matrices for queries, keys, and values in the \(h\)-th head. For instance, multimodal physiological data are collected from multiple sources (electrodes and sensors), thereby forming a spatial network. To identify the physical distribution of the signals, graph convolutional neural networks (GCNs) were introduced, while the GCNs model spatial sensor relationships in graphs with adjacent nodes. GCNs have been widely used in ESD and ESP \cite{wei2024compact}. DL is well-suited to multimodal data due to its ability to handle variability, automate feature extraction, and reduce manual effort. However, it is essential to note that deep learning models are often trained and evaluated on the same data distribution, which may be less suitable for clinical applications, necessitating the use of advanced variants.
\subsubsection{Transfer Learning Methods}
Transfer learning mitigates the scarcity of labeled data and inter-patient variability by fine-tuning pretrained models on multimodal SDP data for ESD and ESP. In practical applications, the initial layers of the neural network, which capture low-level features such as signal rhythms and peak values, are kept fixed. Subsequent layers are fine-tuned to adapt to the target dataset. The optimization procedures are given in the literature \cite{liu2025uncertainty},\cite{zhao2025generalizable}. For instance, domain adaptation has been widely used in ESD and ESP. Another study applied UDA for EEG-based cross-subject prediction, improving multi-subject adaptability \cite{zong2024self}. The mathematical expressions are as follows \cite{zong2024self}.
\begin{equation}
\mathcal{L} = \mathcal{L}_{\text{class}} + \lambda \mathcal{L}_{\text{domain}}
\end{equation}
where \(\mathcal{L}\) is the total loss function; \(\mathcal{L}_{\text{class}}\) is the classification loss (e.g., cross-entropy for seizure states); \(\lambda\) is a weighting hyperparameter balancing the losses; and \(\mathcal{L}_{\text{domain}}\) is the domain adaptation loss (e.g., discrepancy between source and target domains). Domain adaptation effectively mitigates negative transfer and enables the development of calibration-free multimodal systems, aligning with emerging paradigms in intelligent epilepsy monitoring. Furthermore, to enhance the accurate and efficient identification of ictal, interictal, and preictal patterns, generative adversarial networks (GANs) have been employed to synthesize realistic synthetic data, alleviate class imbalances, and bolster overall model robustness \cite{abou2024generative}. Due to challenges in collecting comprehensive physiological data, transfer learning offers a practical approach for seizure detection and prediction.
\subsubsection{Adaptive Learning Methods}
Adaptive methods dynamically update parameters for non-stationary signals and patient variability \cite{afzal2025mt}. Supervised adaptation uses labeled data; the learning rules are as follows:
\begin{equation}
\theta_{t+1} = \theta_t - \eta \nabla_{\theta} \left[ \mathcal{L}(\theta_t; x_t, y_t) + \lambda \mathcal{R}(\theta_t, \theta_{t-1}) \right]
\end{equation}
where \(\theta_{t+1}\) and \(\theta_t\) are the model parameters at time steps \(t+1\) and \(t\), respectively; \(\eta\) is the learning rate; \(\nabla_{\theta}\) denotes the gradient with respect to \(\theta\); \(\mathcal{L}(\theta_t; x_t, y_t)\) is the loss function computed on input \(x_t\) and label \(y_t\); \(\lambda\) is a regularization weight; and \(\mathcal{R}(\theta_t, \theta_{t-1})\) is a regularization term (e.g., parameter difference) to ensure smooth updates. The explanations of the equations are given in the literature \cite{afzal2025mt} while semi-supervised incorporates unlabeled data, expressed as:
\begin{equation}
\mathcal{L} = \mathcal{L}_{\text{sup}}(\theta; x_l, y_l) + \lambda \mathcal{L}_{\text{unsup}}(\theta; x_u)
\end{equation}
and the consistency loss:
\begin{equation}
\mathcal{L}_{\text{unsup}} = | f(\theta; x_u) - f(\theta; \tilde{x_u}) |^2
\end{equation}
where \(\mathcal{L}\) is the total loss; \(\mathcal{L}_{\text{sup}}(\theta; x_l, y_l)\) is the supervised loss on labeled data \(x_l\) and labels \(y_l\); \(\lambda\) is the weighting hyperparameter; \(\mathcal{L}_{\text{unsup}}(\theta; x_u)\) is the unsupervised loss on unlabeled data \(x_u\); \(f(\theta; \cdot)\) is the model output; and \(\tilde{x_u}\) is an augmented version of \(x_u\). For instance, teacher-student frameworks achieve AUC 94 to 96\% with 50\% fewer labels \cite{baghersalimi2024m2skd}. Unsupervised adaptation employs contrastive learning. The mathematical representations are given as follows:
\begin{equation}
\mathcal{L} = -\sum \log \frac{\exp(\text{sim}(z_i, z_j^+)/\tau)}{\sum_{k \neq i} \exp(\text{sim}(z_i, z_k)/\tau)}
\end{equation}
where \(\mathcal{L}\) is the contrastive loss; \(\text{sim}(\cdot, \cdot)\) is a similarity function (e.g., cosine similarity); \(z_i\), \(z_j^+\) are embeddings of a positive pair (e.g., augmented versions of the same sample); \(z_k\) are embeddings of other samples (negatives); \(\tau\) is the temperature parameter; and the summation is over positive and negative pairs in a batch. Self-supervised pretraining with variational autoencoders and clustering enhances stability. Although sensitivity to initialization presents a challenge, contrastive pretraining improves both stability and generalization. The detailed explanation of the advanced decoding methods is given in Table \ref{tab-03}.

\begin{table}[htbp]
\centering
\caption{Summary of Advanced Neural Decoding Methods for Seizure Detection and Prediction}
\label{tab-03}
\setlength{\tabcolsep}{6pt}
\renewcommand{\arraystretch}{1.2}
\begin{tabular}{p{2cm} p{1.5cm} p{4cm}}
\toprule
\textbf{Literature} & \textbf{Decoding Method} & \textbf{Algorithm} \\ \bottomrule
\cite{sr108}--\cite{wei2024compact} & Deep Learning & CNN, GCN, LSTM, Attention Mechanisms, Transformers, Hybrid Models \\
\cite{zhao2025generalizable}--\cite{abou2024generative} & Transfer Learning & Domain Adaptation, Few-Shot Learning, Domain Adversarial, Fine-Tuning \\
\cite{afzal2025mt}--\cite{qi2025epileptic} & Adaptive Learning & Unsupervised, Semi-Supervised, Supervised Learning \\
\bottomrule
\end{tabular}
\end{table}

\subsection{Evaluation Metrics for Multimodal Epileptic Seizure Detection and Prediction}
The clinical deployment of AMSDP in clinical settings, integrating physiological signals, imaging data, and video monitoring, requires stringent performance evaluation, with metrics from single-modality systems adaptable to these approaches. In ESD and ESP literature, key metrics including sensitivity, specificity, accuracy, precision, false-positive rate, and AUC-ROC are commonly utilized \cite{8626042} \cite{ahmad2024attention},\cite{ahmad2023hybrid}.In practice, predictors identify periods with a high probability of seizure occurrence. Key performance metrics include the seizure occurrence period, the time window during which a seizure may occur, and the seizure prediction horizon, the interval between the alarm and the start of the seizure occurrence period \cite{ahmad2023hybrid} . The area under the ROC curve is also used to evaluate the performance of ES prediction algorithms \cite{ahmad2024robust}.

\subsection{Edge Computing and Real-Time Monitoring}

\begin{figure}[!ht]
\centering
\includegraphics[width=\linewidth]{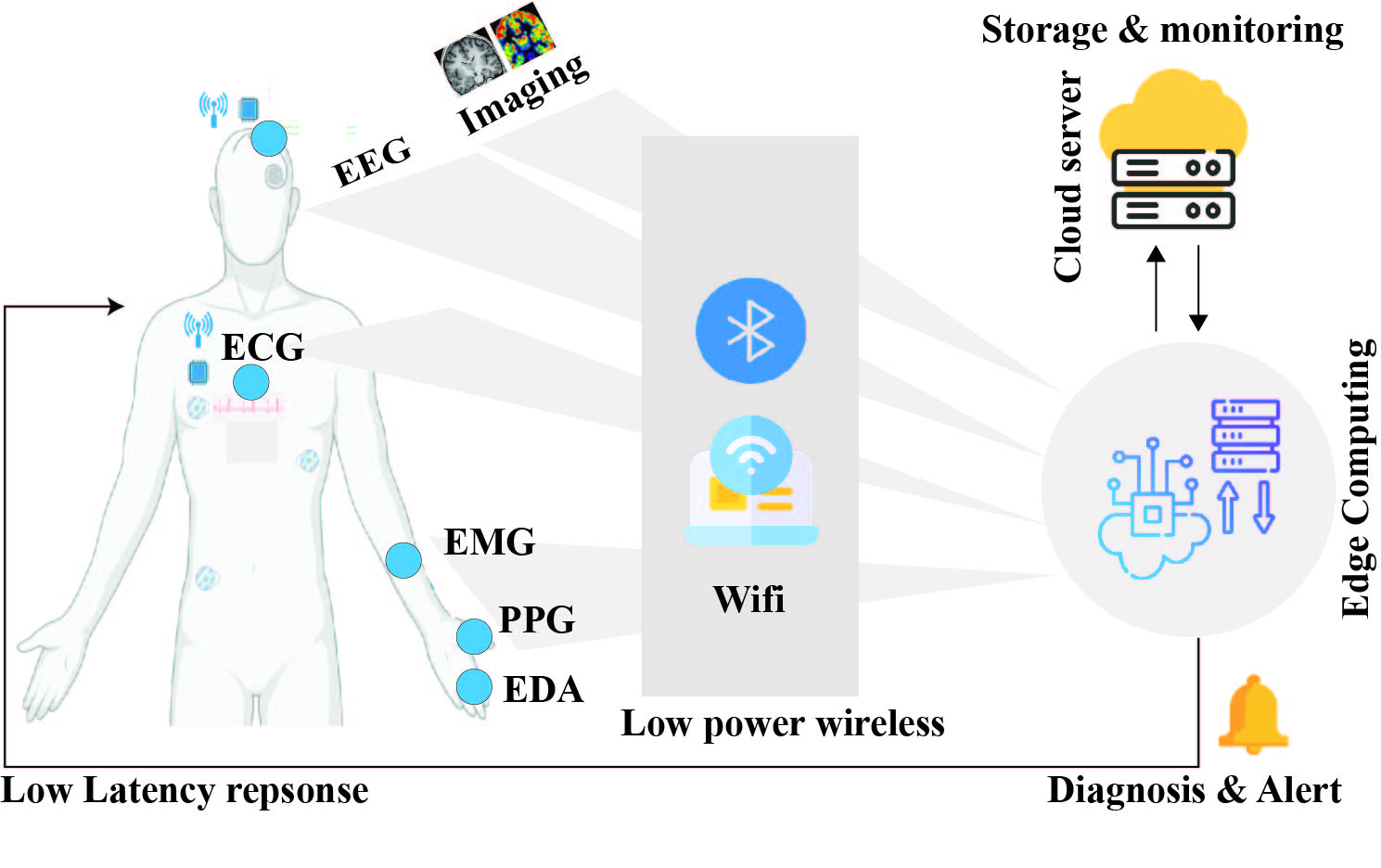}
\caption{Multimodal edge computing framework utilizing heterogeneous physiological signals for seizure diagnosis and treatment.}
\label{fig-6}
\end{figure}

Edge computing plays a critical role in real-time seizure detection and prediction by reducing latency and operational costs through local data processing \cite{idrees2022edge}, while managing high-volume multimodal streams, supporting real-time alerts and neurostimulation, and enhancing privacy. Architectures include sensing layers with BLE, Wi-Fi, or LTE-M connectivity \cite{baek2025edge}. Edge devices perform denoising, feature extraction, and quantized inference. Cloud integration provides storage, analytics, and federated learning. This is followed by on-device inference with quantized deep learning models for seizure detection and prediction. The edge gateway can securely notify clinicians and family members via mobile devices to facilitate rapid, context-aware decision-making, as shown in the Fig. \ref{fig-6}. In addition to edge capabilities, cloud computing offers scalable, long-term storage, cohort-level analytics, model orchestration, and retraining through methods such as federated or continual learning. It also supports compliance-grade audit trails for data governance.

\begin{table*}[htbp]
\centering
\caption{Advanced Fusion Approaches in Multimodal Epileptic Seizure Detection and Prediction}
\label{tab-07}
\begin{tabular}{p{3.8cm} p{3.2cm} p{2.8cm} p{3.5cm} p{3.2cm}}
\toprule
\textbf{Reference} & \textbf{Modality Combination} & \textbf{Fusion Level} & \textbf{Methods} & \textbf{Task} \\
\midrule
Vandecasteele et al. \cite{vandecasteele2021power} & EEG + ECG & Data & Concatenation; Bandpass Filtering; STFT & Seizure detection  \\
Bi et al. \cite{bi2022multi} & EEG + Autonomic & Data & Feature Stacking; 1D-CNN & Seizure detection  \\
Yuan et al. \cite{yuan2018multi} & EEG + ECG & Decision & LSTM; Correlation Scoring; Ensemble & Seizure detection \\
Al-Qazzaz et al. \cite{al2024automatic} & EEG + ECG/EMG & Decision & ViewNet; Majority Voting &  Seizure detection \\
Aung et al. \cite{zhu2024afsleepnet} & EEG + HRV & Feature & Fuzzy Entropy; SVM & Seizure detection \\
Wu et al. \cite{luo2023dual} & EEG + ECG & Feature & Self-Attention; Gating; MLP & Seizure prediction  \\
Zhang et al. \cite{zhou2024learning} & EEG + Video & Feature & Dual Attention; GCN & Seizure detection \\
\bottomrule
\end{tabular}
\end{table*}
\section{Future Directions and Research Opportunities}
\subsection{Curse of Multimodal Data Dimensionality}
The development of efficient AMSDP  systems has intensified the challenge of data dimensionality. Integrating diverse data sources, including physiological signals, imaging modalities, and video-based monitoring, results in high-dimensional feature spaces. This increased dimensionality complicates the analysis and interpretation of multimodal data, increases computational costs, and heightens the risk of model overfitting \cite{yang2022multimodal}.
Physiological signals often include multiple sensor channels, imaging modalities produce large volumes of voxel- or pixel-based features, and video streams provide extensive spatiotemporal data. Advanced dimensionality reduction and feature selection are essential for extracting relevant information from each modality and minimizing redundancy. Techniques such as principal component analysis, independent component analysis, and deep learning-based feature embedding are commonly used to reduce dimensionality across these data types \cite{irani2025time}. Developing adaptive algorithms that dynamically select the most informative features or channels from each data source is essential for timely and robust seizure detection/prediction. Balancing the preservation of critical multimodal information with the reduction of data complexity is necessary to ensure that models generalize effectively and operate efficiently in clinical settings. Future research should focus on optimizing multimodal fusion strategies and advancing dimensionality reduction techniques specifically designed for heterogeneous seizure detection and prediction datasets.
\subsection{High-quality Multimodal Data and Self-supervised Learning}
Developing innovative multimodal systems for clinical applications requires high-quality, well-annotated multimodal datasets. However, due to data privacy regulations and ethical considerations, the collection and sharing of such data remain significant challenges, particularly the complexity and heterogeneity of clinical modalities (e.g., neuroimaging, electrophysiological recordings). High-quality multimodal datasets are essential for training advanced deep learning models, including foundation models and vision-based models \cite{khan2025comprehensive}. To address the privacy concern inherent in medical data, federated learning offers a promising solution by enabling model training across distributed clinical datasets while preserving patient confidentiality and data sovereignty \cite{baykara2025federated}.
Furthermore, supervised learning approaches remain effective when sufficient labeled data are available, but the scarcity of annotated multimodal datasets limits their scalability \cite{zong2024self}. Therefore, self-supervised learning (SSL) techniques have demonstrated potential by leveraging large volumes of unlabeled clinical data for clinical applications such as epileptic seizure detection and prediction. Advancements in SSL frameworks can be interesting work in multimodal seizure detection and  prediction, thereby enhancing the generalizability and robustness of AI-driven systems in clinical settings.
\subsection{Real-Time Prognosis of Epileptic Seizure Patient }
Efficient real-time MSDP require low latency, optimal hardware performance, and minimal signal noise. Achieving these objectives necessitates advanced filtering and modeling techniques. For example, deep learning methods provide high accuracy but demand significant computational resources, restricting real-time implementation to high-performance hardware \cite{ahmad2023hybrid}. In contrast, edge computing reduces latency but may compromise sensitivity. Recent strategies combine nonlinear dynamics, deep learning, and signal processing to improve accuracy and latency, yet no consensus has emerged regarding a single optimal solution \cite{alnaseri2024review}. Further advancements are required, including optimized deep learning architectures, enhanced feature extraction, and effective trade-offs between accuracy and latency. Advanced machine learning and deep learning models demonstrate significant diagnostic potential, with some studies reporting 100\% accuracy \cite{ahmad2024robust}. However, existing research has not comprehensively addressed the clinical requirements necessary for successful translation into practice. Algorithm performance varies across acquisition devices and sensors \cite{zhu2021multimodal}. For clinical adoption, algorithms must demonstrate robustness across multiple modalities, sensor types, and patient demographics \cite{oh2025multi}. Additionally, they must operate efficiently in real time and integrate into comprehensive workflows that account for latency constraints, alarm management, and clinician involvement, rather than relying solely on simulation-based performance metrics.
Wearable monitoring devices enhance accessibility and patient adherence in ambulatory settings but encounter challenges, including reduced signal quality due to fewer electrodes, unstable contact, and motion artifacts. System limitations, including battery life and restricted on-device processing capabilities, further affect reliability. Future research should focus on advanced filtering techniques and the development of adaptive algorithms to improve the accuracy of seizure detection and prediction while minimizing false alarms.

\section{Conclusion}
\label{sec6}
In this survey paper, we thoroughly investigate the concept of advanced multimodal seizure detection and prediction, addressing the limitations of traditional unimodal EEG-based systems. This topic is extensive and has gained significant attention in recent years. Our findings indicate that most existing surveys concentrate primarily on signal modalities, whereas our survey highlights the evolution of seizure detection and prediction technologies beyond just single modalities. We examined multiple modalities, including physiological signals, neuroimaging, and video, along with the core challenges associated with seizure detection and prediction systems that do not rely solely on EEG data. The AMSDP system integrates advanced pipelines that include advanced fusion techniques, feature engineering, neural coding with clinical validation, and edge computing for ESD and ESP. Our survey addresses a crucial gap in the current literature and serves as a valuable resource for researchers and practitioners in the field. The ultimate goal is to advance neurotechnology toward wearable, multimodal solutions for effective epilepsy monitoring.

\section*{References}
\bibliographystyle{IEEEtran}
\bibliography{sun_ref}

 \end{document}